\documentclass{rashidi}

\usepackage{bm}              

\title{\LARGE \bf
Predicting the Price of Gold in the Financial Markets Using Hybrid Models
}

\author{Mohammadhossein Rashidi$^{1}$, Mohammad Modarres$^{1}$}
\email{m.rashidi@sharif.edu}

\begin{document}

\maketitle
\thispagestyle{firstpage}  
\pagestyle{empty}

\begin{abstract}

Predicting the price that has the least error and can provide the best and highest accuracy has been one of the most challenging issues and one of the most critical concerns among capital market activists and researchers. Therefore, a model that can solve problems and provide results with high accuracy is one of the topics of interest among researchers. In this project, using time series prediction models such as ARIMA to estimate the price, variables, and indicators related to technical analysis show the behavior of traders involved in involving psychological factors for the model. By linking all of these variables to stepwise regression, we identify the best variables influencing the prediction of the variable. Finally, we enter the selected variables as inputs to the artificial neural network. In other words, we want to call this whole prediction process the "ARIMA– Stepwise Regression - Neural Network" model and try to predict the price of gold in international financial markets. This approach is expected to be able to be used to predict the types of stocks, commodities, currency pairs, financial market indicators, and other items used in local and international financial markets. Moreover, a comparison between the results of this method and time series methods is also expressed. Finally, based on the results, it can be seen that the resulting hybrid model has the highest accuracy compared to the time series method, regression, and stepwise regression.\par

\end{abstract}

\section{INTRODUCTION}

From ancient times, many have attempted to predict prices using econometric methods and techniques available in this field. With the passage of time and the increasing universality of access to international financial markets, the need for better prediction has been felt more than ever before. This is because decision-making for investment in financial markets is heavily dependent on predicting expected ranges and asset fluctuations. This sense and need for modeling and predicting fluctuations in financial time series, especially in optimizing portfolios, identifying market characteristics, and asset valuation, is more evident than other cases. Also, in order to reduce risk for investors, accurate prediction of the price and future components of the portfolio has become more necessary than ever before.\par

Over time, with the increasing importance of financial markets worldwide, researchers have developed and introduced innovative methods. Jan Tinbergen can be called the architect of the first econometric model. After him, Donald Cochrane and Guy Orcutt were able to develop the previous models to a valuable conclusion about the use of econometric models. They showed that if there was a positive correlation between the remaining parameters of a model, it would reduce the variance of the estimated parameters and, as a result, the F and t-test values would be overstated. It was at this time that James Durbin and Geoffrey Watson were able to introduce a test to find this correlation. It seemed that this problem had been solved, but this did not continue much until 1970, as this problem had a major developmental obstacle in empirical econometrics. [1]\par
This problem persisted until James Box and William Jenkins published a book on time series forecasting and analysis, which gained a lot of attention. They were able to provide a simple way to predict future variables by defining a non-homogeneous model. Today we know this method as "Box-Jenkins analysis". [1]

Continuing on this path, in 1982 and by an Englishman, the conditional heteroscedasticity variance model, called autoregressive conditional heteroscedasticity (ARCH), was defined[2]. A year later, in 1986, Bolerslev expanded and improved this model, which resulted in the development of the GARCH model according to his studies[3]. Many researchers have used these models and their families in time series prediction studies, including Nels Callis, Ken Fsy, and Kroner in 1995, who used GARCH and standard deviation to predict daily fluctuations in cocoa, corn, cotton, gold, and silver prices[4]. Continuing these efforts and in 2012, Trouk and Liang compared GARCH, ARCH, TGARCH, and TARCH models in examining gold price fluctuations, and their research resulted in the superiority of the TGARCH model over other models. As a result, we can mention prediction and simulation, which are two important applications of GARCH in financial markets[5].\par
One of the observations regarding non-linear time series has been that the GARH models did not have the ability to respond to these types of patterns, which were complex models, and did not yield satisfactory results. Therefore, researchers turned to the use of artificial intelligence and used this tool for a long time. The available tools provided to researchers by artificial intelligence include genetic algorithms, fuzzy networks, and artificial neural networks, which were one of the most suitable of these tools, and many researchers have used this method in their research. In 2004, Hamid and Eqbal used a hybrid GARH-artificial neural network model to study the fluctuations of the S\&P 500 index in Istanbul stock in 2009[6][7]. In 2008, Parizi and Diaz predicted the price of gold using an improved old neural network technique and two newer techniques in artificial networks[8]. Continuing in 2011, Hajizadeh, Safi, and colleagues predicted fluctuations in the S\&P 500 index using a hybrid model[9]. In 2014, Kritstianpoulr and Minotolu predicted fluctuations in the price of gold using a hybrid neural network-GARH model and using data related to volatility indices and the Dow Jones index, as well as the euro-dollar and yen-dollar currency pairs, along with oil prices[10]. In another study that used an ARIMA time series model to predict the price of gold, the accuracy of the model developed by Jianwei and colleagues was found to be 96.2\%[11]. The result of all these studies was an improvement in predictions. For example, in a study by Kritstianpoulr and Minotolu on predicting fluctuations in the price of gold using a hybrid neural network-GARH model, the results showed a 25\% improvement in the instantaneous gold price and a 38\% improvement in the future gold price, which was an important achievement achieved by using ANN. \par
In this research, the goal is to obtain a model that can have the least error and provide reliable predictions of future gold prices. \par
This research consists of three chapters. In the next chapter, we introduce the problem. After that, we provide relevant results. Finally, last chapter, conclusions about this research will be discussed.\par
\section{Problem Description}
The model used is a hybrid model that includes a combination of time series methods, technical analysis variables, multi-step regression, and artificial neural networks. In this study, three categories of data are used: gold, oil, and the euro-dollar currency pair. Gold is a variable that we want to obtain its value in the future, and oil and the euro-dollar currency pair are variables that, along with other variables such as technical analysis variables, help us achieve this goal. In this context, the data related to gold in this modeling is equal to the final, first, highest, and lowest daily gold prices traded in the global financial market. The time interval of the data is from January 1, 2015, to January 1, 2019. The data from January 1, 2015, to January 1, 2018, are related to test data, and the data from January 2, 2018, to the beginning of January 2019 are related to measurement data. The desired data is obtained from the com.investing website, and the prices related to gold are equal to their value in US dollars.

The eight sets of data related to oil in this modeling are equal to the final price of US heavy oil, which is traded daily in the US financial market. The time interval of the data is from the beginning of 2015, i.e., January 1, 2015, to January 1, 2019. The data from January 1, 2015, to January 1, 2018, are related to test data, and the data from January 2, 2018, to the beginning of January 2019 are related to measurement data. \par

\subsection{Time Series}
{ Time series models are one of the important branches of econometrics. They are one of the oldest and fundamental methods used by researchers and traders in financial markets to predict future prices. Analyzing and examining time series helps identify relationships and patterns in observations, which enables us to define rules and laws for existing variables and to predict the desired variable in the future.\par
In statistics and econometrics, especially in the analysis of time series, a stationary autocorrelated integerated moving average, or ARIMA(p,d, q), and a more comprehensive model, ARMA(p,q), are used to better understand the model or predict the future in time series analysis. These models are used where the data is non-stationary. In this case, we can disappear the non-stationarity of the data with one or two differentiation (corresponding to first-order differencing), and the possibility of estimating the data in new observations arises.\par
In this model, p, d, and q are real non-negative numbers that determine the degree of autocorrelation, one-persistence, and moving average in time series analysis. ARMA models are an important part of the Box-Jenkins approach to time series modeling. If one of these values is equal to zero, it is usually written as AR, I, or MA.\par
Hence, we can write an (q, d, p) ARIMA model as:
\begin{equation*}
\resizebox{1\linewidth}{!}{$(1 - \phi_1 B - \cdots - \phi_p B^p)(1 - B)^d y_t = \mu + (1 + \theta_1 B + \cdots + \theta_q B^q) \varepsilon_t$}
\end{equation*}
where $B$ is the backshift operator, $\mu$ is the constant, $\phi_i$ and $\theta_i$ are the $p$ and $q$ autoregressive and moving average coefficients, respectively, and $d$ is the degree of differencing.
\subsection{Technical Analysis Indicators}

One of the most important tools that traders and researchers in the financial world use is the knowledge and skills related to technical analysis. This tool, which is the result of mathematical relationships, is actually an indicator of specific relationships between prices, trading volumes, and the behavior of players, etc. Technical analysis is a means by which trends are identified and, in addition to that, a prediction of what will happen in the future can be achieved. In this article, by reviewing other sources, five categories of the most important technical analysis indicators are examined. These indicators are:\par

1. Exponential Moving Average Indicator\par 
2. Relative Strength Indicator\par
3. Stochastic Oscillator Indicator \%K\par
4. Stochastic Oscillator Indicator \%D\par
5. Williams Indicator\par
\subsubsection{Moving Average Indicator}
One of the most popular and widely used indicators is the exponential moving average, which represents the average value of prices over arbitrary time periods. This indicator places more weight on recent price data points.  Its equation is defined as:\par
\[
\text{EMA} = (P \cdot \frac{2}{n + 1}) + \text{EMA}_{\text{previous}}
\]
Where:
$P$ is the current price
$n$ is the number of periods being analyzed
$\text{EMA}_{\text{previous}}$ is the previous EMA value.
\subsubsection{Relative Strength Indicator}
To calculate the Relative Strength Index (RSI), the following steps are taken. In order to use the Relative Strength Index, it is common to analyze the index over a period of 14 days. However, it is possible to analyze it over other periods as well. But what is customary in financial markets among traders and researchers is to use the number 14:\par

1. The daily price changes are obtained. If the price change is positive, it is recorded in the "gains" column, and if it is negative, it is recorded in the "losses" column.\par
2. The average gains or losses are calculated for the selected time period, which was previously 14 days. The first usable value is available after 14 days.\par
3. Finally, using the formula shown below, the value of this index is obtained for the selected time period:
\begin{equation*}
RSI=100-\frac{1}{1+\frac{\sum\limits_{i=0}^{n-1}DW_i}{\sum\limits_{i=0}^{n-1}UP_i}}\times 100
\end{equation*}
In which, $UP_i$ is the upward price change for period i, $DW_i$ is the downward price change for period i, and n is the number of periods being analyzed.
\subsubsection{Stochastic Oscillator Indicator \%K, \%D}
The stochastic oscillator examines the existing correlation between the closing price of a security and its natural range over a specific time period. This oscillator is composed of two lines, where the main line is called \%K and the second line is called \%D, which is equal to the moving average of \%K. The following equations are used for the stochastic oscillator:
\begin{equation*}
   \%K = \frac{C_t - L_{t-n}}{HH_{t-n} - LL_{t-n}}\times100
\end{equation*}
where $C_t$ is the latest price, $L_{t-n}$ is the lowest price over the period, $HH_{t-n}$ is the highest high over the past t days, and $LL_{t-n}$ is the lowest low over the past t days.
\begin{equation*}
   \%D = \frac{\sum_{i=0}^{n-1} \%K_{t-i}}{n} 
\end{equation*}

Where \%D is equal to the moving average of \%K, and n is the number of periods in consideration.
\par
\subsubsection{Williams Indicator}
This indicator is like a torque that measures extreme buying or selling levels and its interpretation and analysis has a lot of similarity with a random oscillator. The difference between them lies in the sharp and smooth internal fluctuations, so that for this indicator, these fluctuations are uniform and smooth.
\begin{equation*}
   \%R = \frac{HH_{t-n} - C_t}{HH_{t-n} - LL_{t-n}}\times100
\end{equation*}

\subsection{Machine Learning Algorithms}
\subsubsection{Stepwise Regression}
The method of operation in this algorithm is as follows: After selecting the desired algorithms, the best one is selected in the first stage. In the second stage, a variable is selected from the remaining variables that, first, has no correlation with the variable selected in the first stage and, second, prevents overestimation and underestimation problems. In the next stage, a variable must be selected that, in addition to not having the problem of overestimation and underestimation, has no correlation with the variables selected in the previous stages. This is done iteratively, which means that it may be observed in one stage that a variable that existed in the initial stages of the model should be removed from the model because new variables that have entered the problem have reduced its impact when combined with them. After each stage, the model accuracy is evaluated with test data, and this back-and-forth process continues until the model accuracy reaches the desired level and there are no excessive selection problems.\par
In this research, after all the desired variables are calculated and selected with this method. we select the best of them and use them as inputs for our artificial neural network.
\subsubsection{Artificial Neural Network}

\par

A neural network can be considered one of the most famous algorithms in machine learning. This algorithm is used in many fields and various types of applications. A neural network can be defined as a combination of artificial neurons, where each neuron performs a certain operation on the input data and produces an output. There are different methods to obtain predictions using random data that is fed into a sigmoid function. Of course, the output of this function can be fed into another sigmoid function, and this process can be repeated.\par
A neural network with three main layers is comprised of an input layer, a hidden layer, and an output layer. The hidden layer itself consists of a number of neurons whose output values are input into another function called the activation function. In the end, the output of the activation functions, which are connected to the last layer with certain weights, are entered into it and provide us with the final result, which can be a unique or multiple estimation of the variables of interest. The accuracy and value of the error resulting from the estimation are calculated by a function called the cost function. The optimal number of neurons in each of the hidden layers is determined through various iterations and experiments.

\section{Model}
After describing the problem, we start to use the data and generate the results for our model.
\subsubsection{Predicting Daily Closing Price with ARIMA}
To find a model that can provide an accurate prediction of future prices, we turn to the training data, which cover from January 1, 2015 to the January 1, 2018 A. We use this data to identify a prediction model. At first glance, Fig 1, it may seem that the time series is stationary and lacks a trend. However, this assumption must be tested by drawing ACF and PACF. 
\begin{figure}
\begin{center}
    
\includegraphics[scale=.65]{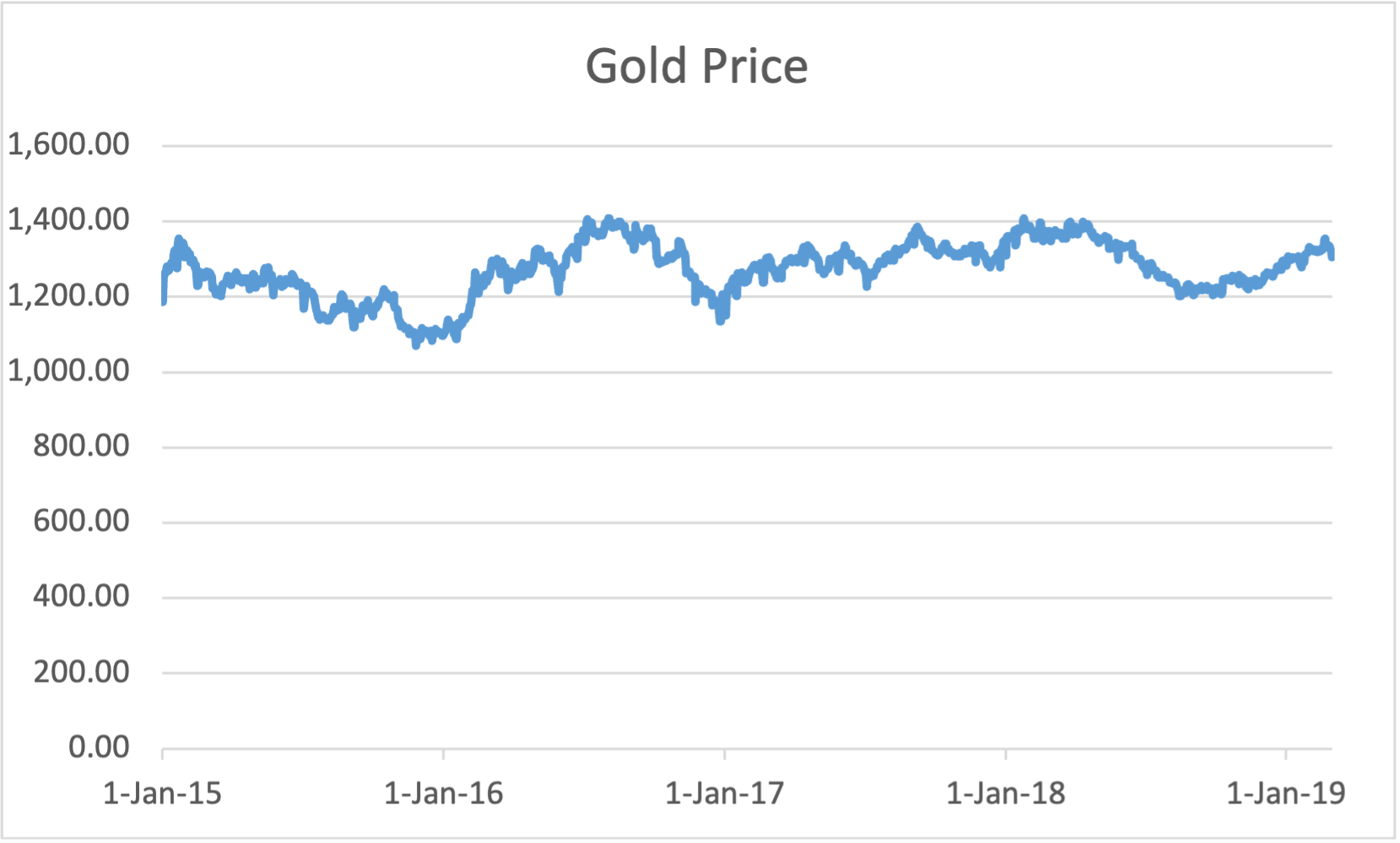}
\caption{The chart of daily closing prices in \$ for gold in a specific time period.}
\end{center}
\end{figure}
As a result, we will plot these two. As we can see in Fig 2, ACF, the graph gradually tends towards zero at different lags, indicating that the time series is not stationary contrary to our initial assumption. Furthermore, we observe that the value of the lag at the first log is also very close to one, which confirms the non-stationarity phenomenon even more. Therefore, based on the observed ACF, we conclude that the mentioned time series is non-stationary.
\begin{figure}
\begin{center}
    
\includegraphics[scale=.6]{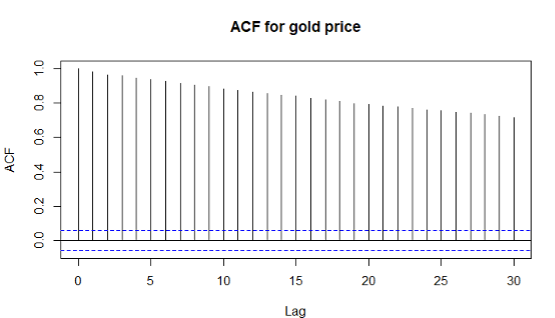}
\caption{ACF for trained data of Gold daily prices.}
\end{center}
\end{figure}
\begin{figure}
\begin{center}
    
\includegraphics[scale=.62]{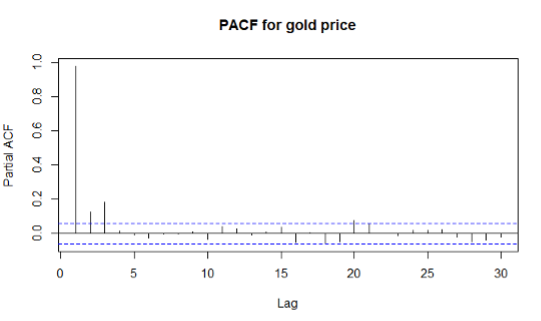}
\caption{PACF for trained data of Gold daily prices.}
\end{center}
\end{figure}
\par
Now, by taking a single difference, Fig 4, we want to investigate whether the mentioned time series becomes stationary or not. In other words, we want to find out whether taking a single difference turns the time series into a stationary one or not. 
\begin{figure}
\begin{center}
    
\includegraphics[scale=.6]{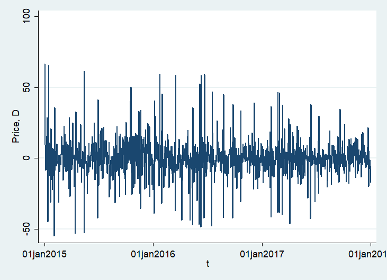}
\caption{One difference in the closing price of gold for training data.}
\end{center}
\end{figure}

\par 
Based on the obtained graph, we can confidently conclude that the resulting plot after taking a single difference is a stationary plot. Therefore, we can proceed to draw its ACF and PACF.
\begin{figure}
\begin{center}
    
\includegraphics[scale=.6]{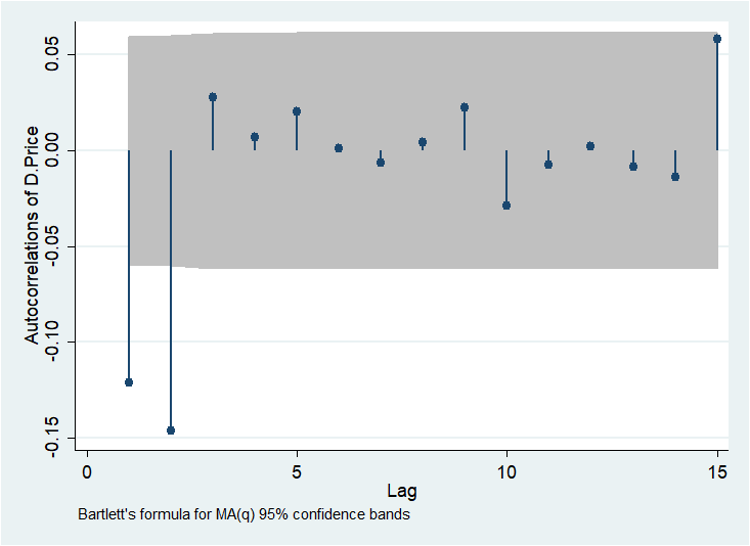}
\caption{ACF for one difference of Gold daily prices.}
\end{center}
\end{figure}
\begin{figure}
\begin{center}
    
\includegraphics[scale=.6]{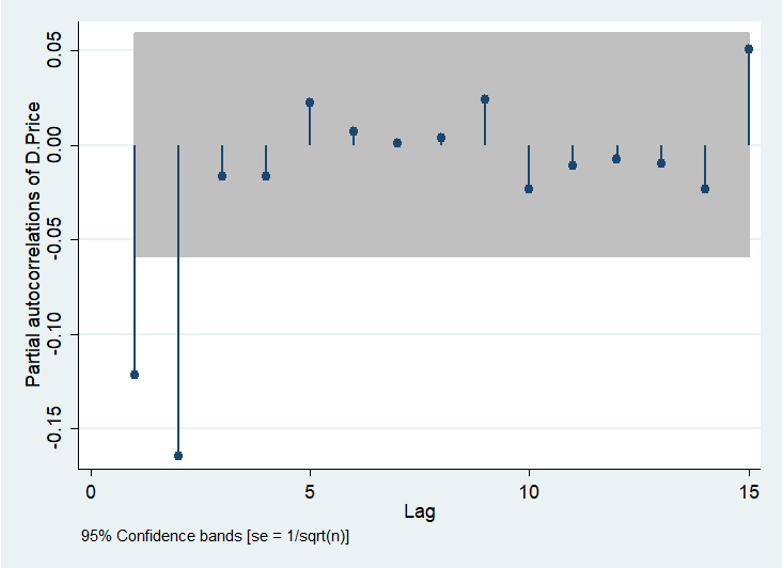}
\caption{PACF for one difference of Gold daily prices.}
\end{center}
\end{figure}
\par
We observe that after taking a single difference, the ACF plot no longer has the gradual slope towards zero that it had before. Instead, it exhibits a cut-off towards zero. Additionally, the values of the lag in the first and second differences are much smaller than zero, which indicates that the series has become stationary. Therefore, we can conclude that in this case, the data has exponentially decreased and moved towards zero.\par

In this case, we can say that the best value for q in (q, MA) is 2, since after taking the first difference and re-plotting the ACF plot, we see that the data has become stationary and we observe a cut-off after the second lag.\par
We want to see at which lag and with what lag value a suitable model for determining the number of p in AR prediction can be obtained. Both AIC and the Schwartz Information Criterion (SIC) determine the second lag as the best value for obtaining the model. Also, in the PACF chart drawn for the lags, we observed a cut-off at the second lag. In this case, we set 2 for p. Moreover, we found that, there is no correlation in the remaining lags of the residuals. Therefore, we can confidently conclude that the obtained model is correct, and there is no correlation among the residuals of this model. In other words, the residuals are white noise. Therefore, we can say that ARIMA(2,1,2) is the best forecasting for this dataset. We found the accuracy for test set is 94.2\% with MAPE method, where MAPE defines with this equation:\par

\begin{equation*}
    MAPE = \frac{1}{n}\times \sum\limits_{i=1}^n \left\vert \frac{A_i - F_i}{A_i} \right\vert \times 100%
\end{equation*}
where $A_i$ is the actual value and $F_i$ is the forecasted value for a particular data point, and $n$ is the total number of data points. The equation calculates the average absolute percentage error between the actual and forecasted values, expressed as a percentage.

\subsubsection{Forecast related to the West Texas Intermediate crude oil index}
The training data for this index was not stationary. As a result, we used one differences data, where d=1 for them. By doing this strategy, the dataset came stationary, as we can see in Fig 7. After that, to find p and q, we used ACF and PACF for that,

\begin{figure}
\begin{center}
    
\includegraphics[scale=.65]{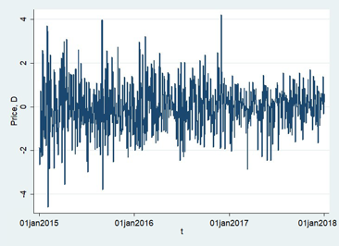}
\caption{One difference of WTI Oil daily prices.}
\end{center}
\end{figure}

To confirm this claim, we should use ACF. As we can see in Fig 8, we can say that based on the smoothed data has been confirmed, firstly because after the first lag, we witnessed a cutoff and in the following interruptions, the data went towards zero, and secondly, the value of the first lag was much smaller than 1. Now, in order to estimate the value of q in MA(q), two values can be considered, each of which must be examined and evaluated separately. In the first assumption, the value is zero, and in the second assumption, the value is equal to 1.
\begin{figure}
\begin{center}
    
\includegraphics[scale=.65]{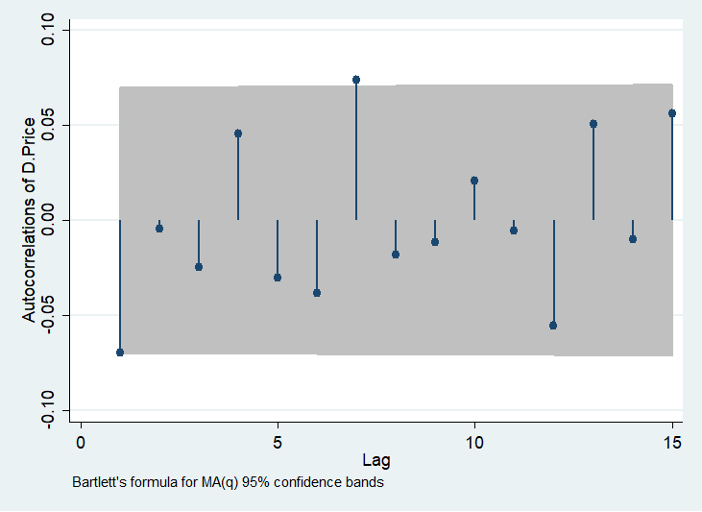}
\caption{ACF for one difference of WTI Oil daily prices train set data.}
\end{center}
\end{figure}

According to the Schwartz test, the value for q should be zero, and based on the AIC test, this value should be set to 1. Since the results based on the Schwartz test are more accurate and take precedence over the result of the AIC test, the value of zero is considered for q.\par
To determine the value of p in AR(p), we turn to the plot related to PACF. It can be seen that after the cutoff, the data goes towards zero in the first interruption, and the value related to the first cutoff is also less than 1. So, two scenarios can be considered for the value of p, one equal to zero and the other equal to one. In this case, we need to examine the (1,1,0) ARIMA model and the (0,1,0) ARIMA model, and whichever has better results for the test data will be selected as the preferred model for predicting oil prices.

\begin{figure}
\begin{center}
    
\includegraphics[scale=.65]{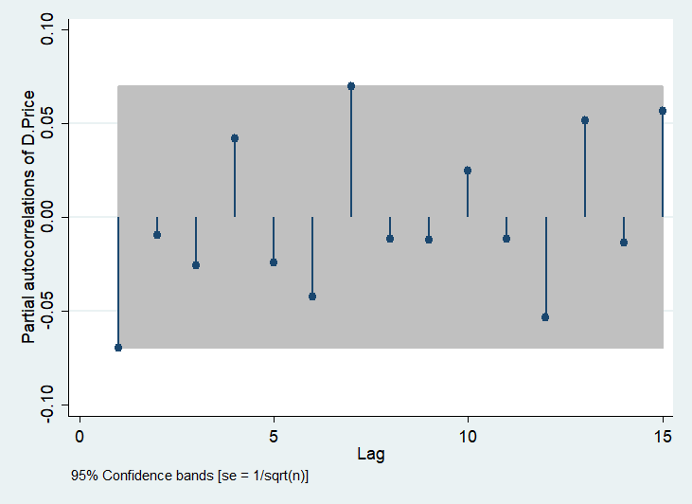}
\caption{PACF for one difference of WTI Oil daily prices for train set data.}
\end{center}
\end{figure}
In this case, the accuracy of the ARIMA(1,1,0)  model, 96.8\%, was higher than the ARIMA(0,1,0) model. Before that, to ensure the validity of the model, we need to investigate the existence of autocorrelation in the residuals in ARIMA(1,1,0), where observed that there is not any autocorrelation in the residuals. 
\subsubsection{Forecasting the Euro-Dollar currency pair} We found that, the training data for this rate was not stationary. So, for the train set dat, we used one differences data, where d=1 for them. By this strategy, we can find that we have stationary dataset. This result is shown in Fig 10.

\begin{figure}
\begin{center}
    
\includegraphics[scale=.65]{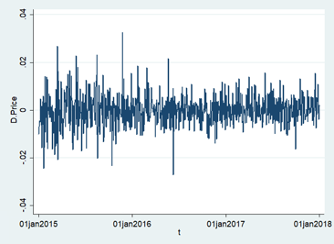}
\caption{One difference of EUR-USD daily rate.}
\end{center}
\end{figure}
Now, we need to use ACf and PACf to find p and q for this dataset. According to what is observed in the corresponding charts in Fig 11, the appropriate value for q in the MA(q) model is equal to zero because the cut-off occurred from the beginning, and in the chart, we see the trend of these two towards zero from that beginning.

\begin{figure}
\begin{center}
    
\includegraphics[scale=.65]{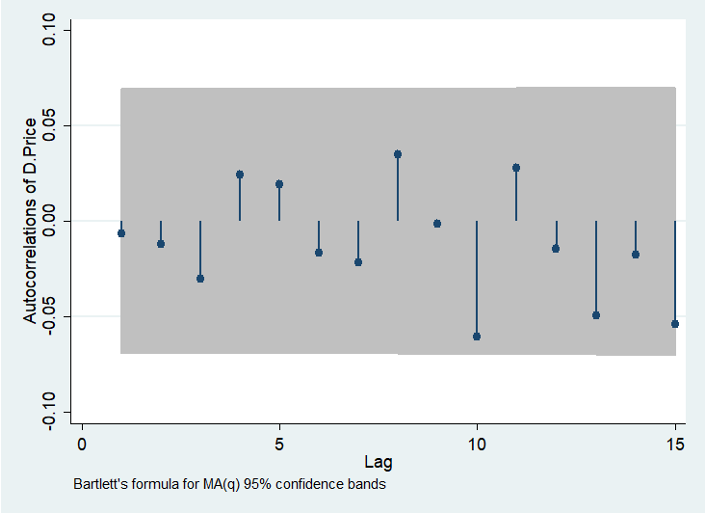}
\caption{ACF for one difference of EUR-USD daily rate in training dataset.}
\end{center}
\end{figure}
This is proved with the results of two tests, Schwartz and AIC. As a result, we set q=0 and we have MA(0) for this part. \par
To obtain the best value for p in the (p)AR model, we now turn to PACF and observe that the cutoff occurred from the beginning. In this circumstances, we have p=0. According to the mentioned results, the model for this dataset is ARIMA(0,1,0). Now, we should find that whether there is autocorrelation in residual values. As the same before, at first, we shouldcalculate the residual values obtained from the estimation equation from ARIMA(0,1,0). Then, based on the obtained values for the residuals, we draw their corresponding ACF and PACF plots given in Fig 12 and Fig 13. 

\begin{figure}
\begin{center}
    
\includegraphics[scale=.73]{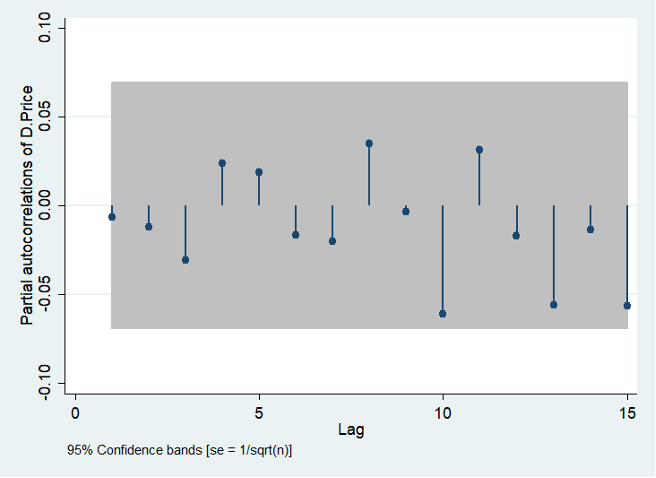}
\caption{ACF for residuals.}
\end{center}
\end{figure}
\begin{figure}
\begin{center}
    
\includegraphics[scale=.73]{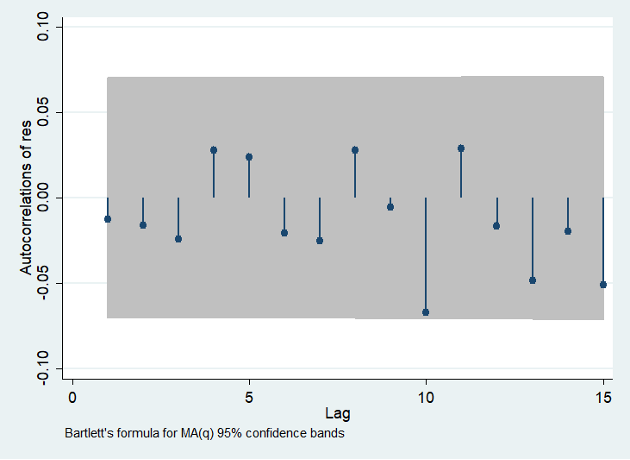}
\caption{PACF for residuals.}
\end{center}
\end{figure}

According to what is observed in the ACF and PACF plots of the residuals, there is no significant relationship among the residuals, and therefore the estimated model is correct, and its accuracy is 97\%.
\subsubsection{Technical Indicators} In this step, we found the technical indicators, EMA(5), EMA(10), RSI, Stochastic K\% and D\%, and Williams for train set. 
\subsubsection{Stepwise Regression}
After obtaining the time series estimation models, we consider technical analysis indicators ,and open price, and high and low price of Gold during each day as independent variables and the closing price of gold as the dependent variable. It is predicted that if we calculate the regression model for the closing price of gold in the future using these data, the model will have a very high error. To prove this claim and to show that the presence of all these dependent variables reduces the quality of the prediction model, we estimate the regression of the gold price using these variables in the learning data range and then proceed with the test data in the corresponding time range using the obtained model.
\par
Therefore, in the first stage, we estimate the regression model in the presence of all variables, which is as follows:

\begin{align*}
y =  0.885 x_1 + 0.022 x_2 - 0.035 x_3 - 0.009 x_4 \\+ 0.115 x_5
- 0.198 x_6 + 0.349 x_7 - 0.106 x_8 \\ + 0.181 x_9 - 0.135
\end{align*}

where $y$ represents the closing price of gold on the next trading day, $x_1$ is the opening price on the current trading day, $x_2$ is the predicted price of the euro-dollar currency pair by a time series model on the next trading day, $x_3$ is the predicted price of oil by a time series model on the next trading day, $x_4$ represents the technical indicator RSI on the current trading day, $x_5$ is the random oscillator indicator K\% on the current trading day, $x_6$ is the random oscillator indicator D\% on the current trading day, $x_7$ is the technical indicator Williams R\% on the current trading day, $x_8$ is the value of the 5-day exponential moving average, and $x_9$ is the value of the 10-day exponential moving average up to the current trading day.\par
The obtained model had low power and high error. By using the MAPE index, the model's power is estimated to be 87\%, which is less than the result for ARIMA forecast. This can be justified by the high number of variables, as having too many independent variables for predicting a value using regression is not always efficient and often reduces the predictive power. Therefore, to improve it, it is necessary to remove some of the independent variables.
\par
Now, in order to demonstrate that this model can be improved with stepwise regression modeling, we turn to modeling. To this end, we use two methods, forward and backward stepwise regression.\par
I) Forward: In this model, at first, none of the variables are included in the model and the first variable to be entered is the one that has the least error according to the Schwartz criterion in estimating the relationships estimated with that variable alone. Thus, the opening price variable on the current trading day is entered into the model. Then, the variables are again entered into the model independently and the variable that has the least error compared to the others according to the Schwartz criterion in the new model containing two independent variables is selected. This process continues until no variable can improve the error by being added to the model. By performing this, the following variables are selected as input variables, and the model formed is the best model with the least error compared to other models.\par
The input variables for the model constructed are the open price, the predicted closing price of oil for the next day, and technical indicators including RSI, stochastic oscillator K\%, Williams technical indicator for the current trading day, and a 10-day exponential moving average up to the current day.
The estimation equations is:
\begin{align*}
    y =  0.436 x_1 - 0.243 x_3 + 0.266 x_4
    + 0.0381 x_5 \\
    + 0.371 x_7 + 0.543 x_9 + 41.664   
\end{align*}
Using the MAPE metric, the model achieves an accuracy of 96.07\%, which supports the fact that using regression models improves the prediction model.\par
II) Backward: As mentioned before, in this method, in the first step, all independent variables are entered into the model, and then one candidate variable is selected as the output variable. The variable that results in the lowest error when removed from the model is selected as the variable to be removed, as its removal results in the model with the lowest error compared to the other outputs.\par
According to the results, the variables selected for the model are the open price, the predicted price of oil by the time series model for the next trading day, RSI technical indicator, Williams technical indicator, and a 10-day moving average. This set has the lowest error compared to other selections, resulting in the following model. The estimated equation is: 
\begin{align*}
    y =  0.432 x_1 - 0.257 x_3 + 0.269 x_4 
    + 0.411 x_7 \\+ 0.548 x_9 + 45.101    
\end{align*}
The accuracy for this model is 97.47\%.\par
In this case, we can conclude that, due to the accuracy, backward strategy is better and we should the extracted independent variables as inputs for our Neural Network.
\subsubsection{Neural Network}
Before starting to execute the neural network, the data must be normalized. After that, we use the variables obtained from stepwise regression as inputs and the corresponding neurons of the input layer in the neural network. In the current model, since the error obtained by the backward method in multi-step regression was less than the forward method, the variables selected by this method, including the opening price, the predicted price of oil for the current trading day, technical indicators RSI, K\% Stochastic, Williams, and EMA(10), are selected as the inputs to the input layer of the neural network. \par
The selected neural network has one hidden layer. The philosophy behind this choice is to improve the accuracy of the neural network and prevent overfitting, as well as increase the speed of calculations. This is because speed is one of the most important advantages for traders in financial markets.\par
To obtain the optimal number of neurons in the hidden layer, values of 1 to 10 neurons were considered as potential candidates for the neural network. The accuracy obtained for each of these selections is explained in the table below, and based on that, the best choice for this neural network is a model with 6 neurons. The Relu activation function was used for this neural network. Moreover, MSE function has been used to measure the error related to neurons and weights of the hidden layer, and MAPE function has been used for neurons and weights related to the output layer. \par

\begin{figure}
\begin{center}
    
 \includegraphics[scale=.56]{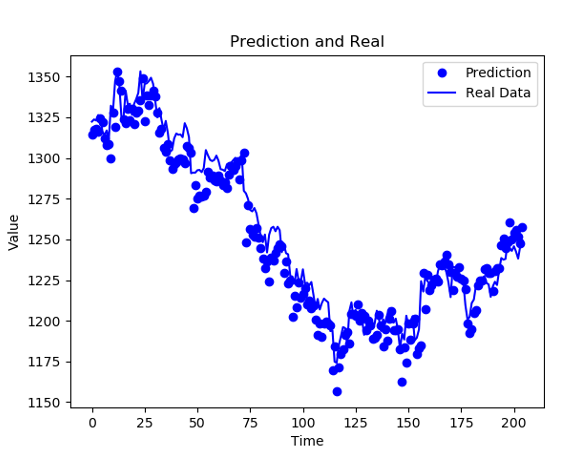}
\caption{Comparison plot of the original data with the best estimate obtained from the artificial neural network.}
\end{center}
\end{figure}

The accuracy of this machine is 99.29\%, which is the highest accuracy among past cases, indicating that this hybrid model has improved the prediction and achieved the highest accuracy and the lowest error compared to other methods. We can see the prediction in Fig 14.

\section{Conclusion}
The existence of many factors that affect the price of an asset forces researchers to go after variables that have the greatest impact on price prediction. Using all variables makes the model execution speed low, which leads to inadequate short-term model response in equilibrium periods, resulting in losses. On the other hand, using too many variables not only slows down model execution, but also affects accuracy. As mentioned in the model, using all variables that are among the most influential variables in predicting prices causes a decrease in accuracy compared to the model obtained by ARIMA. Therefore, we turn to the regression method to improve this issue. By running this model, the model accuracy increased compared to the model obtained by ARIMA and reached 97.47\%.\par

To further improve this error, we turn to the artificial neural network. One of the most important differences in this study compared to other studies is that in other studies, all variables that have a significant impact on prediction are selected as input neurons of the neural network, which slows down model processing speed and reduces accuracy. Therefore, in this model, variables selected by the regression method are chosen as input to the first layer of the neural network to eliminate this drawback. With the application of this method, it is observed that the accuracy of the model increased and in networks with six neurons in the middle layer, an accuracy of 99.29\% was achieved.\par

According to the results, it can be concluded that the hybrid model {\textbf"ARIMA - Technical - Regression Stages - Artificial Neural Network"} had higher accuracy than other models.\par
It is recommended to use other methods for variable selection in future studies, such as random forest. In stock prediction, fundamental variables can also be included in the model. Additionally, for predictive maintenance, genetic algorithms or other metaheuristic algorithms can be used to implement the model on a machine.

\end{document}